# Monte Carlo Bayesian Reinforcement Learning

Yi Wang                                                                            WANGY@COMP.NUS.EDU.SG
Kok Sung Won                                                                    KOKSUNG@COMP.NUS.EDU.SG
David Hsu                                                                           DYHSU@COMP.NUS.EDU.SG
Wee Sun Lee                                                                        LEEWS@COMP.NUS.EDU.SG
Department of Computer Science, National University of Singapore, Singapore 117417, Singapore

## Abstract

Bayesian reinforcement learning (BRL) encodes prior knowledge of the world in a model and represents uncertainty in model parameters by maintaining a probability distribution over them. This paper presents Monte Carlo BRL (MC-BRL), a simple and general approach to BRL. MC-BRL samples *a priori* a finite set of hypotheses for the model parameter values and forms a discrete partially observable Markov decision process (POMDP) whose state space is a cross product of the state space for the reinforcement learning task and the sampled model parameter space. The POMDP does not require conjugate distributions for belief representation, as earlier works do, and can be solved relatively easily with point-based approximation algorithms. MC-BRL naturally handles both fully and partially observable worlds. Theoretical and experimental results show that the discrete POMDP approximates the underlying BRL task well with guaranteed performance.

## 1. Introduction

A major obstacle in reinforcement learning is slow convergence, requiring many trials to learn an effective policy. Model-based Bayesian reinforcement learning (BRL) provides a principled framework to tackle this difficulty. To speed up convergence, BRL encodes prior knowledge of the world in a model. It explicitly represents uncertainty in model parameters by maintaining a probability distribution over them and chooses actions that maximize the expected long-term reward with respect to this distribution. One approach to BRL is to cast it as a partially observable Markov decision process (POMDP) $\mathcal{P}$ (Duff, 2002). The state of $\mathcal{P}$ is a pair $(s, \theta)$, where $s$ is the discrete world state for the reinforcement learning task and $\theta$ is the unknown continuous model parameter. POMDP policy computation automatically analyzes both aspects of each action: its reward and its contribution towards inferring unknown model parameters, thus achieving optimal trade-off between exploration and exploitation.

Despite its elegance, this approach is not easy to use in practice. Since model parameters are continuous in general, $\mathcal{P}$ has a hybrid state space and requires the restrictive assumption of conjugate distributions to represent beliefs during the policy computation (Duff, 2002; Poupart et al., 2006; Ross et al., 2007; Poupart & Vlassis, 2008).

We propose *Monte Carlo Bayesian Reinforcement Learning* (MC-BRL), a simpler and more general approach to BRL, based on the following observation: although there are infinitely many parameter values, it may be possible to compute an approximately optimal policy without considering all of them, if the objective is good average performance with respect to a prior distribution $b_{\mathcal{P}}^0$ of model parameters. We sample a finite set of values from $b_{\mathcal{P}}^0$ and form a discrete POMDP $\hat{\mathcal{P}}$ whose state is $(s, \hat{\theta})$, with $\hat{\theta}$ taking values from the sampled set only. This discrete POMDP $\hat{\mathcal{P}}$ approximates the hybrid POMDP $\mathcal{P}$. $\hat{\mathcal{P}}$ does not require conjugate distributions for belief representation and can be solved much more easily with existing point-based approximation algorithms, *e.g.*, (Kurniawati et al., 2008). MC-BRL also naturally handles both fully and partially observable worlds.

We show that MC-BRL is approximately Bayes-optimal with a bounded error in the average case. The output-sensitive bound indicates that if a small approximately optimal policy exists, then a small number of samples is sufficient for $\hat{\mathcal{P}}$ to approximate $\mathcal{P}$ well. In other words, if we treat $\mathcal{P}$ as a generalization of $\hat{\mathcal{P}}$ with a richer model parameter space, a small policy results in better generalization. This nicely mirrors similar results in learning theory. We also provide experimental results evaluating MC-BRL on four distinct domains, including one from an application in autonomous vehicle navigation.

## 2. Background

### 2.1. MDP and POMDP

An MDP is a tuple $\langle S, A, T, R, \gamma \rangle$, where $S$ is a set of world states, $A$ is a set of actions, $T(s, a, s')$ specifies the transition probability of reaching state $s'$ when taking action $a$ in state $s$, $R(s, a, s')$ specifies the reward received when taking action $a$ in state $s$ and reaching state $s'$, and $\gamma$ is a discount factor.

A policy $\pi : S \to A$ for an MDP is a function that specifies which action to take in each state $s \in S$. The value of a policy $\pi$ is defined as the expected cumulative discounted reward

$$E\left(\sum_{t=0}^{\infty} \gamma^t R\left(s_t, \pi(s_t), s_{t+1}\right)\right),$$

where the expectation is with respect to the random variable $s_t$, the state at step $t$. The aim of the MDP is to find an optimal policy $\pi^\star$ with maximum value.

MDPs assume that the agent can directly observe the world state. POMDPs generalize MDPs by allowing partially observable states. Formally, a POMDP is a tuple $\langle S, A, O, T, Z, R, \gamma \rangle$, where $S, A, T, R, \gamma$ are as defined in the case of MDP, $O$ is a set of observations, and $Z(s', a, o)$ is the observation function that specifies the probability of observing $o$ when action $a$ was taken in the previous step and the current state is $s'$.

In a POMDP, the agent does not know for sure its state. Instead, it maintains a probability distribution or *belief* $b(s)$ over the state space $S$. A policy $\pi : \mathcal{B} \to A$ for a POMDP is a mapping from the belief space to actions. The value of $\pi$ at a belief $b$ is defined as

$$V_\pi(b) = E\left(\sum_{t=0}^{\infty} \gamma^t R\left(b_t, \pi(b_t), b_{t+1}\right) \mid b_0 = b\right),$$

where the expectation is with respect to the random variable $b_t$, the belief at step $t$. Given an initial belief $b_0$, the aim of the POMDP is to find an optimal policy $\pi^\star$ with maximum value at $b_0$.

### 2.2. Related Works

One common approach to BRL adopts the proposal in (Duff, 2002) and casts BRL as a POMDP $\mathcal{P}$ with a hybrid state space (Wang et al., 2005; Poupart et al., 2006; Ross et al., 2007; Castro & Precup, 2007; Poupart & Vlassis, 2008; Ross & Pineau, 2008). To maintain the posterior belief of continuous model parameters, it requires either a closed-form representation or effective approximate inference techniques. Instead of solving $\mathcal{P}$ directly, MC-BRL approximates it with a discrete POMDP $\hat{\mathcal{P}}$ by sampling from the prior distribution and takes advantage of the recent advances in point-based discrete POMDP algorithms. This way, we avoid the restrictive assumption of close-form belief representation and obtain a simpler and more general approach.

Sampling has been used extensively in BRL (Castro & Precup, 2007; Ross et al., 2007; Poupart & Vlassis, 2008; Ross & Pineau, 2008; Asmuth et al., 2009). However, the earlier works draw samples from the posterior distributions to speed up planning for $\mathcal{P}$ or to maintain beliefs efficiently. This is conceptually different from our approach, which samples hypotheses from the model parameter space *a priori* to form $\hat{\mathcal{P}}$ and works exclusively with the sampled hypotheses afterwards.

Our theoretical result shares a similar idea with that for the (PO)MDP algorithm PEGASUS. (Ng & Jordan, 2000). The PEGASUS analysis bounds the number of samples required to find a good policy in a policy class with finite VC-dimension. Our result does not assume such a policy class. It provides an output-sensitive bound that depends on the size of the policy actually computed, instead of a worst-case bound for all policies in a class.

## 3. Monte Carlo BRL

### 3.1. BRL as POMDP

To simplify the presentation, let us first consider BRL of an MDP. Given an MDP $\langle S, A, T, R, \gamma \rangle$, the task of BRL is to find an optimal policy when the transition function $T$ is unknown. Let $\theta = \{\theta_{sas'} | s, s' \in S, a \in A\}$ denote the collection of unknown parameters of the MDP, where $\theta_{sas'} = T(s, a, s')$. It has been shown that the BRL problem can be formulated as a POMDP $\mathcal{P} = \langle S_\mathcal{P}, A_\mathcal{P}, O_\mathcal{P}, T_\mathcal{P}, Z_\mathcal{P}, R_\mathcal{P}, \gamma, b_\mathcal{P}^0 \rangle$ (Duff, 2002). The state space $S_\mathcal{P} = S \times \Theta$ is the cross product of the MDP states $S$ and the parameter space $\Theta$. A state $(s, \theta)$ consists of a world state $s$ of the MDP and a hypothesized value $\theta$ of the unknown parameter. The actions $A_\mathcal{P}$ are identical to the actions $A$ in the MDP. Assuming the parameter $\theta$ does not change over time, the transition function is defined as

$$\begin{aligned} T_\mathcal{P}(s, \theta, a, s', \theta') &= \Pr(s', \theta' | s, \theta, a) \\ &= \Pr(s' | s, \theta, a, \theta') \Pr(\theta' | s, \theta, a) \\ &= \theta_{sas'} \delta_{\theta\theta'}, \end{aligned}$$

where $\delta_{\theta\theta'}$ is the Kronecker delta that takes value 1 if $\theta = \theta'$ and value 0 otherwise. The observation of the POMDP $\mathcal{P}$ indicates the current MDP state. Therefore, we define $O_\mathcal{P} = S$ and $Z_\mathcal{P}(s', \theta', a, o) = \delta_{s'o}$. The reward does not depend on the parameter $\theta$, so we have $R_\mathcal{P}(s, \theta, a, s', \theta') = R(s, a, s')$. Finally, we put a prior distribution $b_\mathcal{P}^0(\theta)$ over $\theta$, which reflects our initial belief of the unknown parameter.

This formulation explicitly represents the uncertainty in the unknown parameter. The parameter $\theta$ forms a component of the POMDP state, which is partially observable and can be inferred based on the history of the observed MDP state/action pairs. By solving the POMDP $\mathcal{P}$, one plans against both the uncertainty in the dynamics and the uncertainty in the model parameter. An optimal policy for $\mathcal{P}$ thus yields an optimal strategy for action selection that balances exploration with exploitation.

Since the parameter $\theta_{sas'}$ takes continuous value, $\mathcal{P}$ has a hybrid state space. Two difficulties arise as a result. The first is how to efficiently maintain a belief for the continuous state variable. In order to attain a closed-form representation, most existing work assumes a conjugate prior $b^0_{\mathcal{P}}$ over the parameter $\theta$, such as the Dirichlet distribution (Dearden et al., 1999; Duff, 2002; Poupart et al., 2006; Ross et al., 2007; Poupart & Vlassis, 2008). The second difficulty is how to solve the hybrid POMDP $\mathcal{P}$ efficiently. Although several approximate algorithms based on function approximation and online planning have been proposed (Duff, 2002; Poupart et al., 2006; Ross et al., 2007), there is no satisfactory answer in general.

### 3.2. Algorithm

MC-BRL is motivated by the following observation. Although there are infinitely many possible values for the parameter $\theta$, it may be possible to compute an approximately optimal policy without considering all of them. MC-BRL consists of two phases, offline and online. Given a prior distribution $b^0(\theta)$ and a sample size $K$, the offline phase of the algorithm works in three steps.

1. Sample $K$ hypotheses $\left(\hat{\theta}^1, \hat{\theta}^2, \ldots, \hat{\theta}^K\right)$ independently from $b^0(\theta)$.

2. Form a discrete POMDP $\hat{\mathcal{P}} = \langle S_{\hat{\mathcal{P}}}, A_{\hat{\mathcal{P}}}, O_{\hat{\mathcal{P}}}, T_{\hat{\mathcal{P}}}, Z_{\hat{\mathcal{P}}}, R_{\hat{\mathcal{P}}}, \gamma, b^0_{\hat{\mathcal{P}}} \rangle$. The state space is the cross product $S_{\hat{\mathcal{P}}} = S \times \{1, 2, \ldots, K\}$. A state $(s, k)$ consists of an MDP state $s$ and an indicator $k$ of the sampled hypotheses for $\theta$. The actions $A_{\hat{\mathcal{P}}} = A$ and observations $O_{\hat{\mathcal{P}}} = S$ are defined in the same way as in Section 3.1. The transition, observation, and reward functions are defined as $T_{\hat{\mathcal{P}}}(s, k, a, s', k') = \hat{\theta}^k_{sas'} \delta_{kk'}$, $Z_{\hat{\mathcal{P}}}(s', k', a, o) = \delta_{s'o}$, and $R_{\hat{\mathcal{P}}}(s, k, a, s', k') = R(s, a, s')$, respectively. Finally, the initial belief $b^0_{\hat{\mathcal{P}}}(k)$ is defined as the uniform distribution over $\{1, 2, \ldots, K\}$.

3. Solve the POMDP $\hat{\mathcal{P}}$ and output a policy $\hat{\pi}$.

In the online phase, the agent then follows the policy $\hat{\pi}$ to select actions.

MC-BRL sidesteps the two technical obstacles of the existing approach based on the hybrid POMDP $\mathcal{P}$. The discrete POMDP $\hat{\mathcal{P}}$ can be readily solved with point-based approximation algorithms (Pineau et al., 2003; Smith & Simmons, 2005; Kurniawati et al., 2008). There is also no restrictive assumption on the form of the prior distribution $b^0(\theta)$. The only requirement is that it is easy to sample from.

We further note that $\hat{\mathcal{P}}$ falls into the class of *mixed observability MDPs* (MOMDPs). Its state $(s, k)$ has mixed observability. While the second component $k$ is hidden, the first component $s$ is fully observable. It has been shown that MOMDPs admit a compact factored representation of the state space, which can be exploited to speed up POMDP planning (Ong et al., 2010). In this paper, we use SARSOP (Ong et al., 2010) to solve $\hat{\mathcal{P}}$ which readily takes advantage of the MOMDP representation.

MC-BRL takes a prior distribution $b^0(\theta)$ as input. In practice, if we know nothing about the true parameter, we use a non-informative prior such as uniform distribution. When there is prior knowledge about the true parameter, more informative prior can be used to bias the hypotheses towards the ground truth.

### 3.3. Generalization to Partially Observable Environments

MC-BRL can be readily generalized to BRL problems under partially observable environments. Suppose we are given a POMDP $\langle S, A, O, T, Z, R, \gamma \rangle$, and we aim to find an optimal policy when both the transition function $T$ and the observation function $Z$ are unknown. The unknown parameters can be denoted as a pair $(\theta, \psi)$, where $\theta$ is as defined before, while $\psi = \{\psi_{s'ao} | s' \in S, a \in A, o \in O\}$ denotes the observation function and $\psi_{s'ao} = Z(s', a, o)$.

MC-BRL can be naturally adapted to address this problem with two modifications to the offline phase. First, it samples the hypotheses from a joint prior distribution $b^0(\theta, \psi)$ instead of $b^0(\theta)$. Second, the POMDP $\hat{\mathcal{P}}$ is modified by setting $O_{\hat{\mathcal{P}}} = O$ and $Z_{\hat{\mathcal{P}}}(s', k', a, o) = \hat{\psi}^{k'}_{s'ao}$. The modified observation function $Z_{\hat{\mathcal{P}}}$ now incorporates the uncertainty in the unknown parameter $\psi$ of the underlying POMDP.

## 4. Theoretical Analysis

MC-BRL uses the discrete POMDP $\hat{\mathcal{P}}$ to approximate the hybrid POMDP $\mathcal{P}$. To analyze the quality of this approximation, we derive a probably approximately correct (PAC) bound on the regret of MC-BRL's solution, compared with the optimal solution to $\mathcal{P}$.

We assume that a POMDP policy $\pi$ is represented as a *policy graph* $G$, which is a directed graph with labeled nodes and edges. Each node of $G$ is labeled with an action $a \in A$ and has $|O|$ outgoing edges, each labeled with a distinct observation $o \in O$. The size of the policy $\pi$, denoted as $|\pi|$, is the number of nodes in $G$. To execute the policy, the agent first picks a node in $G$ according to the initial belief. It then takes the action associated with the node, receives an observation, and transits to the next node by following the edge labeled with that observation. The process then repeats.

The policy graph representation allows us to establish the correspondence between policies for $\mathcal{P}$ and $\hat{\mathcal{P}}$. If $\pi$ is a policy for $\mathcal{P}$, then it is also a valid policy for $\hat{\mathcal{P}}$, and vice versa, as $\mathcal{P}$ and $\hat{\mathcal{P}}$ share the same action space $A$ and observation space $O$.

Suppose that MC-BRL forms the discrete POMDP $\hat{\mathcal{P}}$ by taking $K$ samples from the initial belief $b_{\mathcal{P}}^0$ of $\mathcal{P}$. There are three policies of interest: an optimal policy $\pi^\star$ for $\mathcal{P}$, an optimal policy $\hat{\pi}^\star$ for $\hat{\mathcal{P}}$, and the policy $\hat{\pi}$ that MC-BRL actually computes. We want to bound the regret of $\hat{\pi}$ against $\pi^\star$. Define $V_\pi$ as the value of a policy $\pi$ for $\mathcal{P}$ with initial belief $b_{\mathcal{P}}^0$, and $\hat{V}_\pi$ as the value of $\pi$ for $\hat{\mathcal{P}}$ with initial belief $b_{\hat{\mathcal{P}}}^0$. The following theorem states our main theoretical result. The proof is given in the supplementary material[1].

**Theorem 1.** *Suppose that $\pi^\star$ is an optimal policy for $\mathcal{P}$ and $\hat{\pi}$ is the policy that MC-BRL computes by taking $K$ samples to form a discrete POMDP $\hat{\mathcal{P}}$. Let $R_{\max} = \max_{s,s' \in S, a \in A} |R(s,a,s')|$. If $\hat{V}_{\hat{\pi}^\star} - \hat{V}_{\hat{\pi}} \leq \delta$, then for any $\tau \in (0,1)$,*

$$V_{\pi^\star} - V_{\hat{\pi}} \leq \frac{2R_{\max}}{1-\gamma} \sqrt{\frac{2((|\hat{\pi}||O|+2)\ln|\hat{\pi}|+|\hat{\pi}|\ln|A|+\ln(4/\tau))}{K}}$$
$$+ \delta$$

*with probability at least $1 - \tau$.*

The theorem says that MC-BRL with a small set of samples produces a good approximate solution $\hat{\pi}$ to $\mathcal{P}$ with high probability, provided that there exists a simple approximate solution $\hat{\pi}$ to $\hat{\mathcal{P}}$. It is interesting to observe that although we formulate and solve the underlying reinforcement learning task as a planning problem, this analysis closely mirrors similar results in learning: if we think of $\mathcal{P}$ as a generalization of $\hat{\mathcal{P}}$ with a richer model parameter space, then the theorem implies that a small policy results in better generalization.

The error bound consists of two terms. The first term decays at the rate $\mathcal{O}(1/\sqrt{K})$. We can reduce it by sampling more hypotheses from the prior, but at the cost of potentially increasing the complexity of the discrete POMDP $\hat{\mathcal{P}}$ and the resulting policy $\hat{\pi}$. The second term $\delta$ bounds the error in the approximate solution to the discrete POMDP $\hat{\mathcal{P}}$. Algorithms such as HSVI (Smith & Simmons, 2005) and SARSOP (Kurniawati et al., 2008) output such bounds as a by-product of POMDP policy computation. We can reduce $\delta$ by running these algorithms longer towards convergence.

---
[1]Available at http://www.comp.nus.edu.sg/~leews/publications/icml2012-supp.pdf.

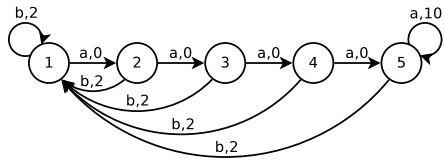

*Figure 1.* Chain problem.

It is also important to observe that the approximate Bayes-optimality of $\hat{\pi}$, quantified by $V_{\hat{\pi}}$, guarantees the average performance of $\hat{\pi}$ with respect to the prior distribution $b_{\mathcal{P}}^0$ of models. It does not guarantee the performance of $\hat{\pi}$ on any particular model.

Our analysis assumes a policy graph representation of POMDP policies. In practice, point-based discrete POMDP algorithms, such as HSVI and SARSOP, typically output policies represented as a set of $\alpha$-vectors, which in principle can be converted to policy graphs.

## 5. Experiments

We now experiment with MC-BRL on both fully observable and partially observable reinforcement learning tasks. First, we evaluate MC-BRL on two small synthetic domains widely used in the existing work on BRL (Sections 5.1 and 5.2). Here the standard setup requires us to measure the performance of an algorithm on particular model parameter values rather than the average performance with respect to a prior distribution of model parameters. Therefore the bound in Theorem 1 is not applicable here. Next, we test MC-BRL on two more realistic domains (Sections 5.3 and 5.4), where we measure the average performance of MC-BRL and show that it performs well in this sense, as our theoretical result guarantees.

All the experiments are conducted on a 16-core Intel Xeon 2.4GHz server.

### 5.1. Chain

We start with the Chain problem used in (Dearden et al., 1998; Poupart et al., 2006). This problem consists of a chain of 5 states and 2 actions $\{a, b\}$. The actions cause the transitions between states and receive corresponding rewards, as shown in Figure 1. The actions are noisy. They slip with probability 0.2 and cause the opposite effect. The optimal policy of this problem is to always perform action $a$.

We consider two versions of the Chain problem. In the

Monte Carlo Bayesian Reinforcement Learning

*Table 1.* Average total rewards (reported with two standard errors) for the Chain problem. The results for Beetle and Exploit are from (Poupart et al., 2006).

|  | Semi-Tied | Full |
|---|---|---|
| MC-BRL ($K = 10$) | 3216±64 | 1661±27 |
| MC-BRL ($K = 100$) | 3603±32 | 1630±25 |
| MC-BRL ($K = 1000$) | 3618±29 | 1646±32 |
| Upper Bound | 3677 | 3677 |
| Beetle | 3648±41 | 1754±42 |
| Exploit | 3257±124 | 3078±49 |
| Q-Learning | 1560±18 | 1560±18 |
| MC-BRL$^+$ ($K = 10$) | – | 3655±24 |
| MC-BRL$^+$ ($K = 100$) | – | 3644±24 |
| MC-BRL$^+$ ($K = 1000$) | – | 3638±24 |

*Table 2.* Average total rewards for the Tiger problem.

|  | Total Reward |
|---|---|
| MC-BRL ($K = 10$) | 68.63±13.45 |
| MC-BRL ($K = 100$) | 113.36±2.38 |
| Upper Bound | 126.34±3.73 |
| Prior Model | 7.12±0.16 |

semi-tied version, we assume that the structure of transitions between states in Figure 1 are given. The only unknown parameters are the 2 slipping probabilities, one for each action. In the full version, we assume that the transition function $T(s, a, s')$ is completely unspecified. This leads to 40 unknown parameters.

We evaluate MC-BRL algorithm using 500 simulations with 1000 steps in each simulation. We test $K = 10, 100,$ and 1000, and use the uniform prior to sample hypotheses. Since it is a stochastic algorithm, we rerun the offline phase of MC-BRL before each simulation, obtain a policy, and then execute that policy online. We run the offline phase up to 180 seconds. The online time is negligible.

Table 1 reports the average (undiscounted) total rewards of MC-BRL. For comparison, we also report an upper bound on the reward that could be achieved only if we had known the true model parameters, as well as the rewards of three alternatives: the Beetle algorithm (Poupart et al., 2006), the Exploit heuristic, which never explores but takes the optimal action with respect to the expected MDP under the current belief, and Q-learning with $\epsilon$-greedy exploration and linear learning rate. For Q-learning, we test a wide range of $\epsilon$ values from 0 to 0.5. The reward for the optimal value is reported.

MC-BRL achieves good performance in the semi-tied version. It obtains near-optimal reward with 1000 samples and is comparable to Beetle. It outperforms Exploit and Q-learning. In the full version, MC-BRL is still better than Q-learning. However, it performs slightly worse than Beetle and is unable to improve the performance substantially with increased number of samples. Exploit performs much better than both MC-BRL and Beetle. However, Exploit relies on a myopic heuristic and does not explore well in general. For example, it performs much more poorly than MC-BRL and Beetle in the semi-tied version.

MC-BRL's performance degrades in the full version, be- cause the sample size is too small to cover the neighborhood of the true parameters within the 40-dimensional parameter space using the uniform prior. To verify this, we conduct another experiment by inserting the true parameter values as one of the samples of MC-BRL. The results, denoted as MC-BRL$^+$ in Table 1, show that MC-BRL achieves good performance in this case. Constructing effective sampling strategies is an important direction for future research.

### 5.2. Tiger

We next test MC-BRL on the Tiger problem (Kaelbling et al., 1998) with partial observability. In this problem, the agent must decide whether to open one of two doors or to listen for the position of the tiger at each time step. Opening the wrong door will cause the agent to be eaten by a tiger with a penalty of $-100$, while opening the correct door will give a reward of 10. Listening costs $-1$ and gives the true position of the tiger with 15% error. We assume that the transition and reward functions are given, but the observation error rates are unknown.

We evaluate MC-BRL using 1000 simulations. Each simulation consists of 100 episodes. In each episode, the agent takes actions and receives observation sequentially. The episode ends when the agent opens a door and the position of the tiger is reset. We test MC-BRL with $K = 10$ and 100. Following (Ross et al., 2007), we use $\texttt{Dirichlet}(3, 5)$ as the prior distribution to sample the unknown parameters. This prior corresponds to an expected error rate 37.5%. We run the offline phase of MC-BRL up to 300 seconds.

Table 2 shows the total reward gained by MC-BRL in 100 episodes, averaged over the 1000 simulations. For reference, we also include the upper bound induced by the true model, and the reward of the prior model in which the observation error rate is set to the prior expectation 37.5%. With $K = 100$, MC-BRL achieves performance close to the upper bound, and is far better than the prior model.

We further look into the evolution of the reward over episodes. Figure 2 shows the reward gained by MC-BRL per episode, averaged over the 1000 simulations. As we do not have the exact settings used in (Ross et al., 2007), we cannot directly compare with their experimental results. However, we can see that MC-BRL quickly learns the un-known parameters and improves over the prior model. It achieves near-optimal performance after about 20 episodes.

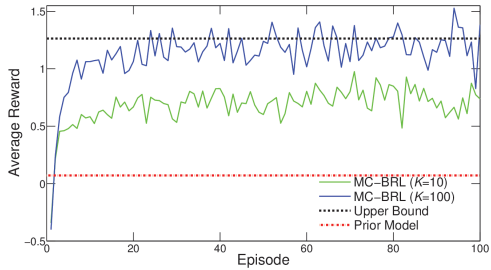

Figure 2. Average reward evolving over episodes for the Tiger problem.

Table 3. Average total rewards over 1000 random opponents for the IPD problem.

|  | Total Reward |
|---|---|
| MC-BRL ($K = 250$) | 917.92±15.97 |
| MC-BRL ($K = 1000$) | 928.03±15.70 |
| Upper Bound | 942.75±15.74 |
| OTFT | 935.80±15.60 |
| Q-Learning | 841.61±13.86 |
| Pavlov | 742.15±15.49 |
| TFT | 661.24±7.98 |
| AP | 520.13±14.87 |

### 5.3. Iterated Prisoner's Dilemma

The Prisoner's Dilemma (Poundstone, 1992) is a well known one-shot two-player game in which each player tries to maximize his own reward by cooperating with or betraying the other. In this section, we studied its repeated version, the Iterated Prisoner's Dilemma (IPD) (Axelrod, 1984), and show that MC-BRL can achieve excellent performance on this problem.

In IPD, the game is played repeatedly and each player knows the history of his opponents moves. A key factor for an agent to gain high reward is the capability to model the opponent's behaviour based on history. It has been shown that any memoryless and one-stage memory opponent can be modeled using 4 parameters $\langle P_S, P_T, P_R, P_P \rangle$, which are the probabilities that the opponent will cooperate in the next step, given the 4 possible situations of the current step: (1) the agent cooperates while the opponent defects (denoted by $S$); (2) the agent defects while the opponent cooperates ($T$); (3) mutual cooperation ($R$); and (4) mutual defection ($P$) (Kraines & Kraines, 1995).

Suppose the agent knows the parameters of its opponent. Then the IPD can be naturally formulated as an MDP. The state of the MDP is the current move of the two players, which takes values from $\{S, T, R, P\}$. The agent needs to select between cooperating or defecting for the next move. The transition function is defined based on the parameters of the opponent. The reward depends on the next state, and is set to $0, 5, 3, 1$ for $S, T, R, P$ respectively, following the setting commonly used in IPD tournaments.

In reality, the parameters of the opponent are unknown. The agent needs to explore the opponent's strategy and at the same time maximize its reward. This leads to a RL problem and we apply MC-BRL to solve it.

We are interested in the average performance of MC-BRL when facing various opponents. Therefore, we randomly select 1000 test opponents by uniformly sampling their parameters. For each opponent, we run the offline phase of MC-BRL for 180 seconds and obtain a policy. We then use the policy to play against the opponent for 300 steps and collect the total reward. This is repeated for 20 times to account for the stochastic behaviour of the opponent. For MC-BRL, we test $K = 250$ and $1000$, and use the uniform prior to sample the parameters. We set the discount factor $\gamma = 0.95$.

Table 3 shows the total rewards averaged over the 1000 opponents. With $K = 250$, MC-BRL already achieves good rewards. With $K = 1000$, it approaches the upper bound, which is achieved by solving the underlying MDP with the true parameters of the opponents.

For reference, we also compare MC-BRL with two classic hand-crafted strategies, Tit-for-Tat (TFT) (Axelrod, 1984) and Pavlov (Nowak & Sigmund, 1993), and the two winning entries of the 2005 IPD tournament, Adaptive Pavlov (AP) (Li, 2007) and Omega Tit-for-Tat (OTFT) (Slany & Kienreich, 2007). These four strategies are used to play against the same 1000 test opponents under the same setting as MC-BRL. The results are summarized in Table 3. MC-BRL achieves comparable reward to OTFT, and significantly outperforms all the others. It is interesting to note that AP, the tournament winner, performs very poorly.

TFT, Pavlov, AP, and OTFT are all specially designed to win the IPD tournaments, while MC-BRL is a general algorithm for BRL and is not optimized for competitions. On the other hand, one should not directly translate the good performance of MC-BRL here to the IPD tournaments, as it is unlikely to face random opponents. However, MC-BRL can use more informative priors to exploit domain knowledge on the opponents, as the other algorithms do.

We further compare MC-BRL with Q-learning. We follow the setting suggested by (Littman & Stone, 2001). The result is shown in Table 3. We can see that MC-BRL significantly outperforms Q-learning on this task.

While MC-BRL achieves good average performance,

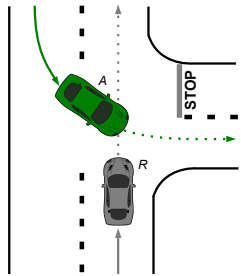

Figure 3. A near-miss accident during the 2007 DARPA Urban Challenge.

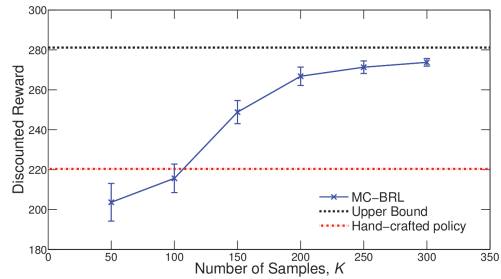

Figure 4. Average discounted total reward for the Intersection Navigation problem versus sample size $K$, reported with two standard error bar.

as our theorem guarantees, it can perform worse than other algorithms when faced with particular opponents. For instance, for the opponent parameterized by $\langle 0.806, 0.108, 0.596, 0.185\rangle$, MC-BRL obtains a much lower reward than that of Q-learning: 596.05 versus 659.1.

### 5.4. Intersection Navigation

This problem is motivated by an accident in the 2007 DARPA Urban Challenge (Leonard et al., 2008). In that event, two autonomous vehicles, $R$ and $A$, approached an uncontrolled traffic intersection as shown in Figure 3. $R$ had the right-of-way and proceeded. However, possibly due to sensor failure or imperfect driving strategy, $A$ did not yield to $R$ and caused a near-miss. This situation is quite common and occurs frequently even with human drivers. Crossing the intersection safely and efficiently without knowing the driving strategy of $A$ poses a significant challenge.

We formulate the problem as a RL problem. The underlying model is a POMDP. The state consists of the positions and velocities of $R$ and $A$. For simplicity, we discretize the environment into a uniform grid. In each step, the agent $R$ can take three actions: accelerate, maintain speed, and decelerate. It then receives an observation on its own state and the state of $A$. Both actions and observations are noisy. The transition function is defined based on the driving strategy of $A$, which is unknown to the agent $R$. The agent receives a reward for crossing the intersection safely, and a large penalty for collision with $A$. A small penalty is given in each step to expedite the agent to cross the intersection faster. Due to space limitation, we give the detailed settings in the supplementary material.

The driving strategy of $A$ is unknown to the agent. We parameterize the driving strategy using 4 parameters: (1) driver imperfection, $\sigma \in [0, 1]$, (2) driver reaction time, $\tau \in [0.5, 2]\,s$, (3) acceleration, $a \in [0.5, 3]\,\text{m/s}^2$, and (4) deceleration, $d \in [-3, -0.5]\,\text{m/s}^2$. A preliminary study shows that this parameterization can cover a variety of drivers such as a reckless driver who never slows down at the intersection and an impatient driver who performs a rolling stop near the intersection. The agent needs to learn the parameters of $A$ and cross the intersection at the same time.

We test MC-BRL on this RL problem. We test a range of $K$ values and sample the parameters from the uniform distribution. Similar to the IPD problem, we are interested in the average performance of MC-BRL with respect to different drivers $A$. Therefore, we uniformly sampled 250 test drivers. For each driver, we run the offline phase of MC-BRL for 1.5 hours and obtain a policy. We then evaluate the policy against that test driver using 200 simulations with 40 steps in each simulation.

Figure 4 shows the average discounted total rewards with discount factor $\gamma = 0.99$. We can see that, as the sample size $K$ increases, the performance of MC-BRL improves quickly. With $K = 300$, it gets close to the upper bound, which is achieved when the true parameters of the driver $A$ are known.

We also compare MC-BRL to a hand-crafted intersection policy that is commonly used in the traffic modeling community (Liu & Ozguner, 2007). With $K = 150$ and above, MC-BRL significantly outperforms that policy. While the hand-crafted policy is not designed to handle noisy observations, we think that the performance gap between the hand-crafted policy and MC-BRL is more likely to be caused by insufficient adaptivity of the hand-crafted policy in learning the driving strategy of $A$.

As a final remark, this problem gives an example where it is more natural to define the prior over the physical properties of the environment. MC-BRL handles such priors easily, although they are challenging to specify using methods that rely on conjugate distributions.

## 6. Conclusion

We have presented MC-BRL, a simple and general approach to Bayesian reinforcement learning. We prove that by sampling a finite set of hypotheses from the model parameter space, MC-BRL generates a discrete POMDP that approximates the underlying BRL problem well with guaranteed performance. We provide experimental results demonstrating strong performance of the approach in practice. Furthermore, MC-BRL naturally handles both fully and partially observable worlds.

One important issue for MC-BRL is to sample the model parameter space effectively. A naive method is to discretize the parameter space uniformly and treat the fixed grid points as samples. This method, however, suffers from the "curse of dimensionality" and is difficult to scale up as the number of parameters increases (Poupart et al., 2006). MC-BRL takes one step further and samples a set of hypotheses independently from a given prior distribution. The promising results obtained in this work open up many possibilities for future investigation, *e.g.*, constructing better informed prior distributions by exploiting domain knowledge and adaptive sampling.

## Acknowledgments

Y. Wang and D. Hsu are supported in part by MoE AcRF grant 2010-T2-2-071 and MDA GAMBIT grant R-252-000-398-490. K.S. Won is supported by an NUS President's Fellowship. W.S. Lee is supported in part by the Air Force Research Laboratory, under agreement number FA2386-12-1-4031. The views and conclusions contained herein are those of the authors and should not be interpreted as necessarily representing the official policies or endorsements, either expressed or implied, of the Air Force Research Laboratory or the U.S. Government.